\pdfoutput=1
\documentclass[final]{cvpr}

\usepackage{times}
\usepackage{epsfig}
\usepackage{graphicx}
\usepackage{amsmath}
\usepackage{amssymb}
\usepackage{lipsum}
\usepackage{booktabs}
\usepackage{bbding}
\usepackage{url}

\usepackage[pagebackref=true,breaklinks=true,colorlinks,bookmarks=false]{hyperref}
\usepackage{color}
\usepackage{enumitem}
\definecolor{mypink3}{cmyk}{0, 0.7808, 0.4429, 0.1412}
\definecolor{myblue2}{cmyk}{0, 0., 0.4412, 0.4808}

\definecolor{myblue}{cmyk}{0, 0.7808, 0., 0.1412}

\def\cvprPaperID{2529} 
\def\confYear{CVPR 2021}

\begin{document}

\title{Semi-supervised Semantic Segmentation with Directional Context-aware Consistency}


\author{Xin Lai$^{1}$\thanks{Equal Contribution}\hspace{1.0cm}Zhuotao Tian$^{1*}$\hspace{1.0cm}Li Jiang$^{1}$\hspace{1.0cm}Shu Liu$^{2}$\\Hengshuang Zhao$^{3}$\hspace{1.0cm}Liwei Wang$^{1}$\hspace{1.0cm}Jiaya Jia$^{1,2}$\\
$^{1}$The Chinese University of Hong Kong~~~
$^{2}$SmartMore~~~
$^{3}$University of Oxford\\
\texttt{\footnotesize
\{xinlai,zttian,lijiang,lwwang,leojia\}@cse.cuhk.edu.hk\quad sliu@smartmore.com\quad hengshuang.zhao@eng.ox.ac.uk
}
}

\maketitle
\thispagestyle{empty}

\begin{abstract}
Semantic segmentation has made tremendous progress in recent years. However, satisfying performance highly depends on a large number of pixel-level annotations. Therefore, in this paper, we focus on the semi-supervised segmentation problem where only a small set of labeled data is provided with a much larger collection of totally unlabeled images. Nevertheless, due to the limited annotations, models may overly rely on the contexts available in the training data, which causes poor generalization to the scenes unseen before. A preferred high-level representation should capture the contextual information while not losing self-awareness. Therefore, we propose to maintain the context-aware consistency between features of the same identity but with different contexts, making the representations robust to the varying environments. Moreover, we present the Directional Contrastive Loss (DC Loss) to accomplish the consistency in a pixel-to-pixel manner, only requiring the feature with lower quality to be aligned towards its counterpart. In addition, to avoid the false-negative samples and filter the uncertain positive samples, we put forward two sampling strategies. Extensive experiments show that our simple yet effective method surpasses current state-of-the-art methods by a large margin and also generalizes well with extra image-level annotations. Our code is available at \url{https://github.com/dvlab-research/Context-Aware-Consistency}.
\end{abstract}

\section{Introduction}

\begin{figure}
\begin{center}

	\centering
    \begin{minipage}  {0.32\linewidth}
        \centering
        \includegraphics [width=1\linewidth,height=0.72\linewidth]
        {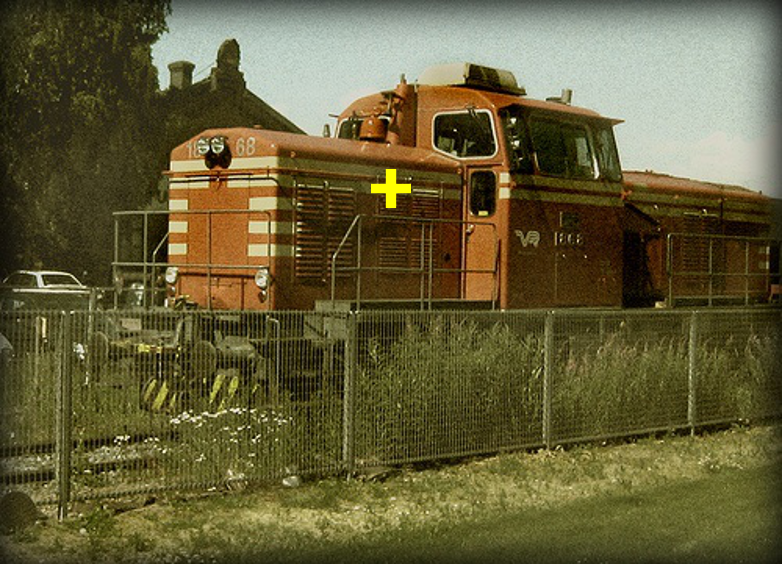}
    \end{minipage}      
    \begin{minipage}  {0.32\linewidth}
        \centering
        \includegraphics [width=1\linewidth,height=0.72\linewidth]
        {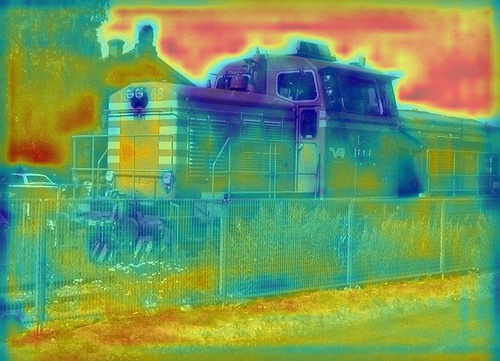}
    \end{minipage}      
     \begin{minipage}  {0.32\linewidth}
        \centering
        \includegraphics [width=1\linewidth,height=0.72\linewidth]
        {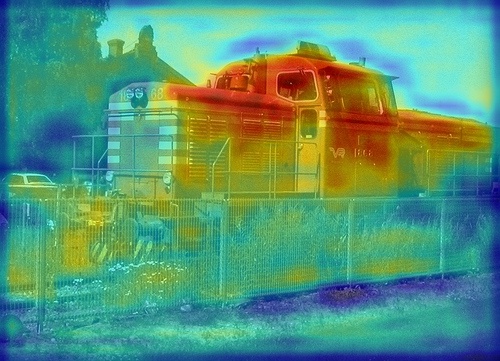}
    \end{minipage} 
	 
	 
    \begin{minipage}[t]{0.32\linewidth}
        \centering
        \includegraphics [width=1\linewidth,height=0.72\linewidth]
        {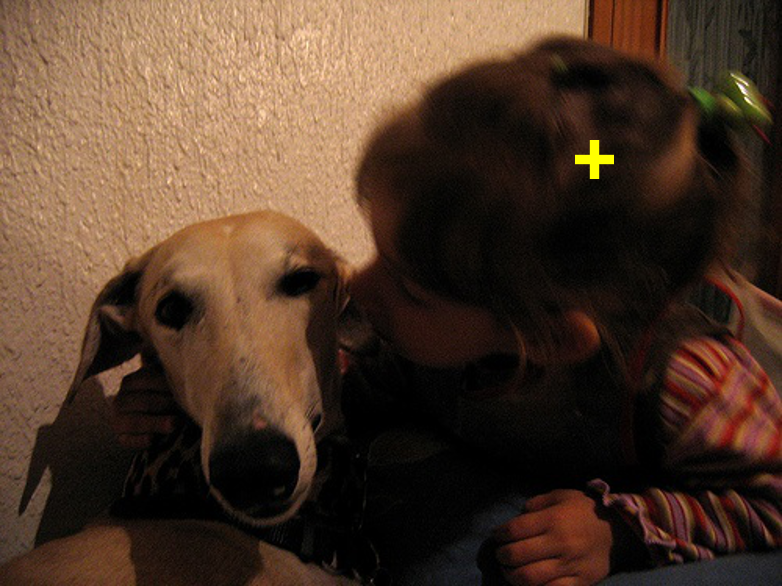} \\\footnotesize Input
    \end{minipage}      
    \begin{minipage}[t]{0.32\linewidth}
        \centering
        \includegraphics [width=1\linewidth,height=0.72\linewidth]
        {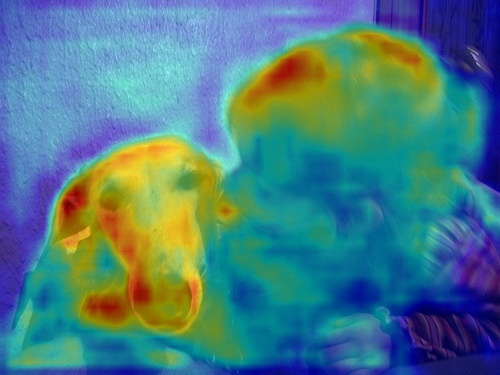} \\\footnotesize SupOnly
    \end{minipage}      
     \begin{minipage}[t]{0.32\linewidth}
        \centering
        \includegraphics [width=1\linewidth,height=0.72\linewidth]
        {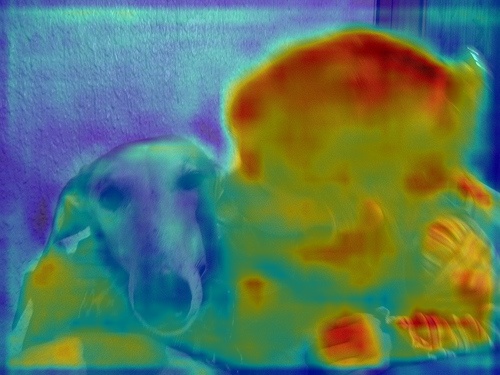} \\\footnotesize Ours
    \end{minipage} 
\end{center}
\vspace{-0.2cm}
\caption{
Grad-CAM~\cite{gradcam} visualizations of the regional contribution to the feature of interest (\ie, the yellow cross shown in the input). The red region corresponds to high contribution.
SupOnly: the model trained with only 1/8 labeled data.  More illustrations are shown in the supplementary.
}
\label{fig:cam}
\vspace{-0.5cm}
\end{figure}

Semantic segmentation, as a fundamental tool, has profited many downstream applications, and deep learning further boosts this area with remarkable progress. However, training a strong segmentation network highly relies on sufficient finely annotated data to yield robust 
representations for input images, and dense pixel-wise labeling is rather time-consuming, \eg, the annotation process costs more than 1.5h on average for a single image in Cityscapes~\cite{cityscapes}. 

To alleviate this problem, weaker forms of segmentation annotation, \eg, bounding boxes~\cite{boxsup,boxsup2}, image-level labels~\cite{imgsup1,imgsup2,imgsup3} and scribbles~\cite{weakscribble1,weakscribble2,weakscribble3}, have been exploited to supplement the limited pixel-wise labeled data. Still, collecting these weak labels requires additional human efforts. Instead, in this paper, we focus on the semi-supervised scenario where the segmentation models are trained with a small set of labeled data and a much larger collection of unlabeled data. 

Segmentation networks can not predict a label for each pixel merely based on its RGB values. Therefore, the contextual information is essential for semantic segmentation. Iconic models (\eg, DeepLab~\cite{deeplabv3+} and PSPNet~\cite{pspnet}) have also shown satisfying performance by adequately aggregating the contextual cues to individual pixels before making final predictions. 
However, in the semi-supervised setting, models are prone to overfit the quite limited training data, which results in poor generalization on the scenes unseen during training. In this case, models are easy to excessively rely on the contexts to make predictions. Empirically, as shown in Fig.~\ref{fig:cam}, we find that after training with only the labeled data, features of \textit{train} and \textit{person} overly focus on the contexts of \textit{sky} and \textit{dog} but overlook themselves. 
Therefore, to prevent the model abusing the contexts and also help enhance self-awareness, our solution in this work is to make the representations more robust to the changing environments, which we call the \textbf{context-aware consistency}.

\begin{figure}
\begin{center}
\includegraphics[width=1.0\linewidth]{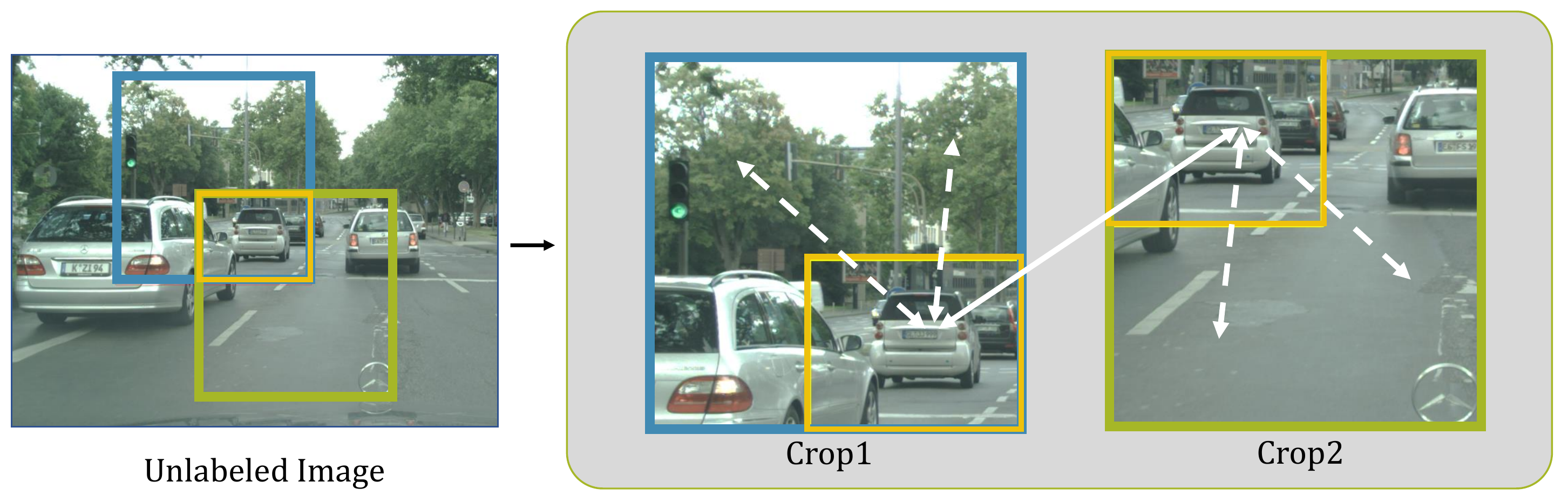}
\end{center}
\vspace{-0.3cm}
\caption{Crop1 and Crop2 are randomly cropped from the same image with an overlapping region. The consistency (represented by the solid white line) is maintained between representations for the overlapping region in the two crops under different contexts (represented by the dashed white line), in a pixel-to-pixel manner. }
\label{fig:intro}
\vspace{-0.3cm}
\end{figure}

Specifically, as shown in Fig.~\ref{fig:intro}, we crop two random patches from an unlabeled image and they are confined to have an overlapping region, which can be deemed that the overlapping region is placed into two different environments, \ie, \textit{contextual augmentation}. Even though the ground-truth labels are unknown, the consistency of high-level features under different environments can still be maintained because there exists a pixel-wise one-to-one relationship between the overlapping regions of the two crops.
To accomplish the consistency, we propose the Directional Contrastive Loss that encourages the feature to align towards the one with generally higher quality, rather than bilaterally in the vanilla contrastive loss.
Also, we put forward two effective sampling strategies that filter out the common false negative samples and the uncertain positive samples respectively. Owing to the context-aware consistency and the carefully designed sampling strategies, the proposed method brings significant performance gain to the baseline.

The proposed method is simple yet effective. Only a few additional parameters are introduced during training and the original model is kept intact for inference, so it can be easily applied to different models without structural constraints. Extensive experiments on PASCAL VOC~\cite{pascalvoc} and Cityscapes~\cite{cityscapes} show the effectiveness of our method. 

In sum, our contributions are three-fold:
\vspace{-0.2cm}
\begin{itemize}
    \item To alleviate the overfitting problem, we propose to maintain context-aware consistency between pixels under different environments to make models robust to the contextual variance.
    \vspace{-0.2cm}
    \item To accomplish the contextual alignment, we design the Directional Contrastive Loss, which applies the contrastive learning in a pixel-wise manner. Also, two effective sampling strategies are proposed to further improve performance.
    \vspace{-0.2cm}
    \item Extensive experiments demonstrate that our proposed model surpasses current state-of-the-art methods by a large margin. Moreover, our method can be extended to the setting with extra image-level annotations.
\end{itemize}


\section{Related Work}
\paragraph{Semantic Segmentation}
Semantic segmentation is a fundamental yet rather challenging task. High-level semantic features are used to make predictions for each pixel. FCN~\cite{fcn} is the first semantic segmentation network to replace the last fully-connected layer in a classification network by convolution layers. As the final outputs of FCN are smaller than the input images, methods based on encoder-decoder structures~\cite{deconvnet,segnet,unet} are demonstrated to be effective by refining the outputs step by step. Although the semantic information has been encoded in the high-level output features, it cannot well capture the long-range relationships. Therefore, dilated convolution~\cite{deeplab,dilation}, global pooling~\cite{parsenet}, pyramid pooling~\cite{pspnet,icnet,denseaspp} and attention mechanism~\cite{danet,ccnet,psanet,asymmetric_nonlocal} are used to better aggregate the contexts. 
Despite the success of these models, they all need sufficient pixel-wise annotations to accomplish representation learning, which costs lots of human effort.

\vspace{-0.3cm}
\paragraph{Semi-Supervised Learning}
Semi-supervised learning aims to exploit unlabeled data to further improve the representation learning given limited labeled data~\cite{entropymin,pseudolabel,semimetric,semigraph}. Adversarial based methods~\cite{cls_gan1,cls_gan2,cls_gan3} leverage discriminators to align the distributions of labeled and unlabeled data in the embedding space. Our method in this paper follows another line based on consistency. VAT~\cite{semivat} applies adversarial perturbations to the output and  $\Pi$-Model~\cite{temporal_ens} applies different data augmentations and dropout to form the perturbed samples and aligns between them. Dual Student~\cite{dualstudent} generates perturbed outputs for the same input via two networks with different initializations. Data interpolation is another feasible way to get perturbed samples in MixMatch~\cite{mixmatch} and ReMixMatch~\cite{remixmatch}. Besides, consistency training can be accomplished with confident target samples. Temporal Model~\cite{temporal_ens} ensembles the predictions over epochs as the targets and makes the outputs consistent with them. Mean Teacher~\cite{meanteacher} yields the target samples via exponential moving average. Also, ideas of self-supervised learning have been exploited to tackle the semi-supervised learning recently~\cite{semiselfsup,S4L}, and we also incorporate the contrastive loss that has been well studied in the self-supervised learning~\cite{contrast1,moco,mocov2,sup_contrast,simclr,simclrv2} as the constraint to accomplish consistency training.

\vspace{-0.3cm}
\paragraph{Semi-Supervised Semantic Segmentation}
Pixel-wise labelling is more costly than image-level annotations. Weak labels including bounding boxes~\cite{boxsup,boxsup2}, image-level labels~\cite{imgsup1,imgsup2,imgsup3,imgsup4} and scribbles~\cite{weakscribble1,weakscribble2,weakscribble3} are used to alleviate this issue, but they still require human efforts. 
To exploit the unlabeled data, adversarial learning and consistency training are leveraged for semi-supervised segmentation. Concretely, both AdvSemiSeg~\cite{AdvSemiSeg} and S4GAN~\cite{s4gan_mittal} utilize a discriminator to provide additional supervision to unlabeled samples. Similar to Mean Teacher, S4GAN~\cite{s4gan_mittal} also uses the teacher-student framework and the final multi-class classifier to filter out uncertain categories by scaling the predictions. \cite{advsemi} adds new samples that are synthesized based on the unlabeled data. The idea of self-correction has been exploited in ECS~\cite{errorcorrect} and GCT~\cite{gct} by creating the Correction Network and Flaw Detector respectively to amend the defects in predictions. Nevertheless, CCT~\cite{CCT} aligns the outputs of the main decoder and several auxiliary decoders with different perturbations to enforce a consistency that improves feature representations. Unlike these methods, our proposed context-aware consistency brings significant performance gain by explicitly alleviating the contextual bias caused by limited training samples.
  
\section{Method}

\begin{figure}
	\centering
    \begin{minipage}  {0.32\linewidth}
        \centering
        \includegraphics [width=1\linewidth,height=1\linewidth] 
        {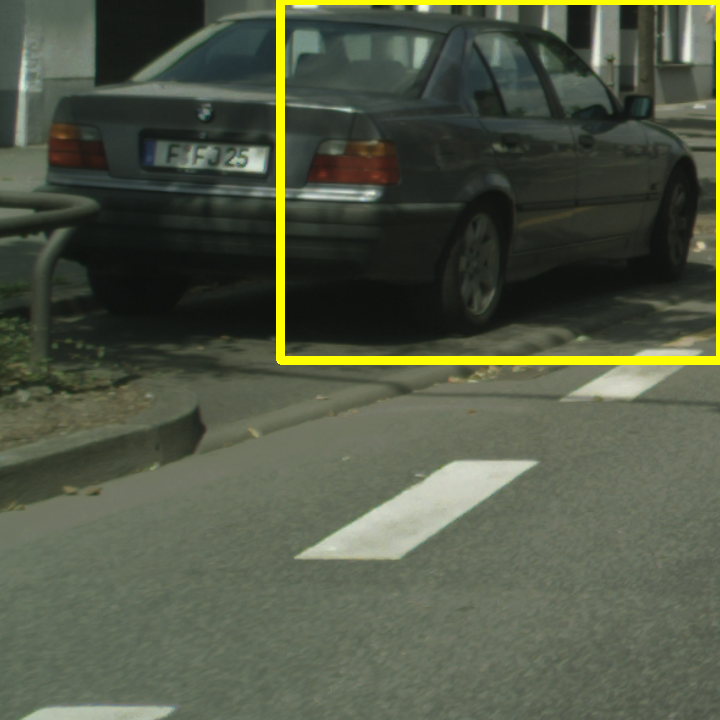}
    \end{minipage}      
    \begin{minipage}  {0.32\linewidth}
        \centering
        \includegraphics [width=1\linewidth,height=1\linewidth] 
        {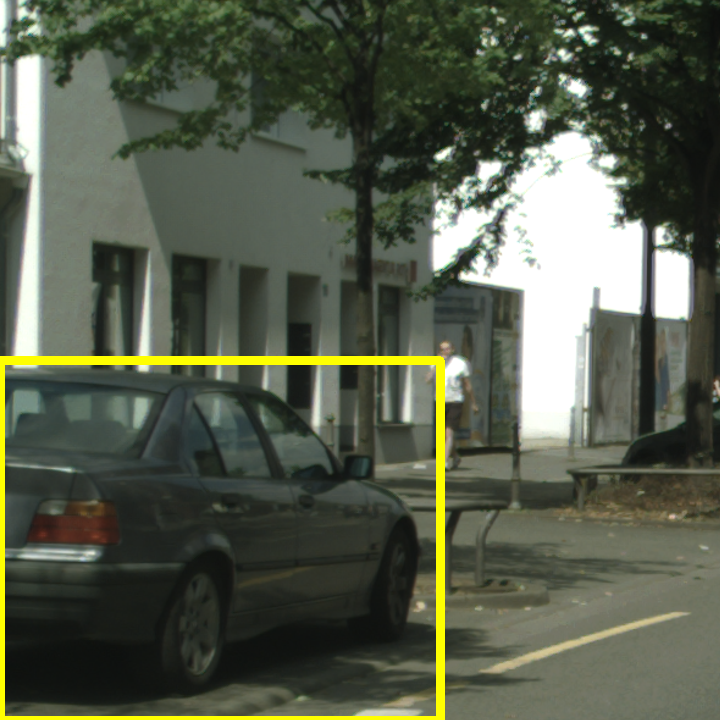}
    \end{minipage}      
     \begin{minipage}  {0.32\linewidth}
        \centering
        \includegraphics [width=1\linewidth,height=1\linewidth] 
        {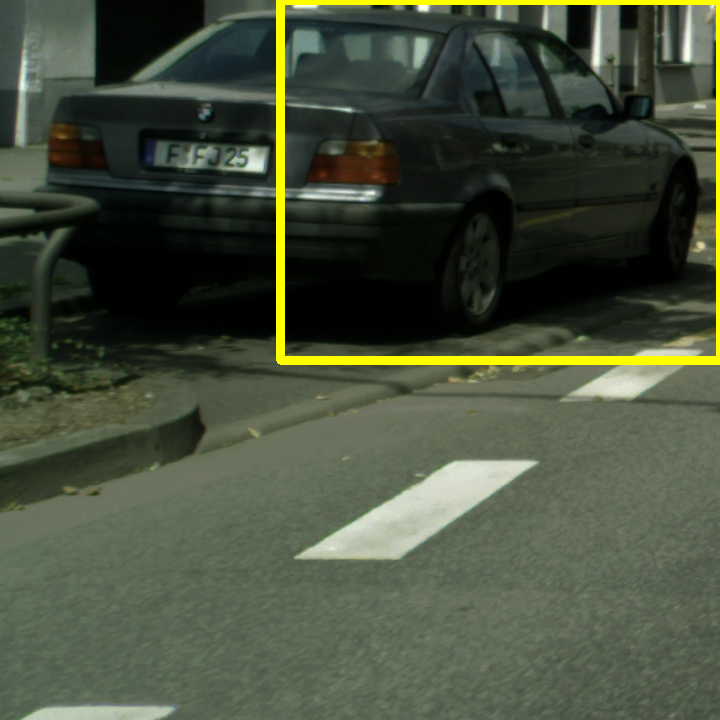}
    \end{minipage} 
	 
	 \vspace{0.1cm}
	 
    \begin{minipage}  {0.32\linewidth}
        \centering
        \includegraphics [width=0.8\linewidth,height=0.8\linewidth] 
        {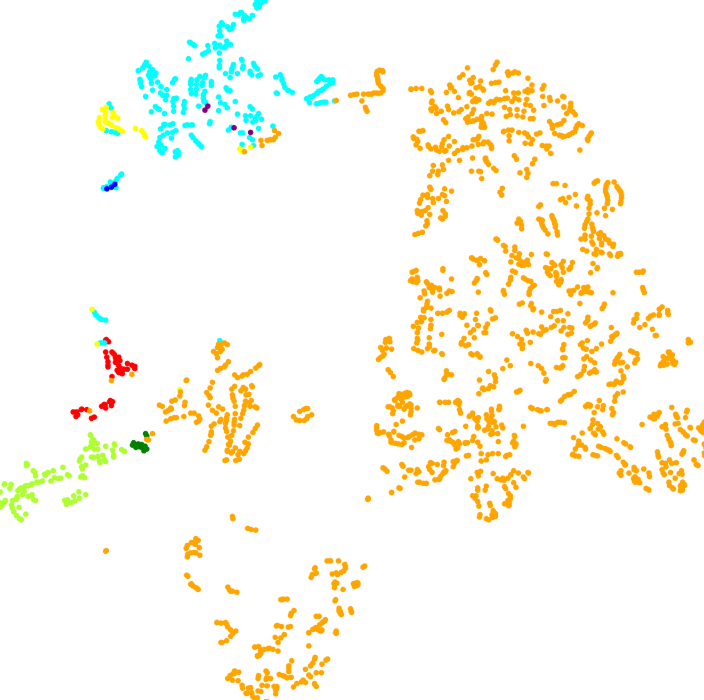}
    \end{minipage}      
    \begin{minipage}  {0.32\linewidth}
        \centering
        \includegraphics [width=0.8\linewidth,height=0.8\linewidth] 
        {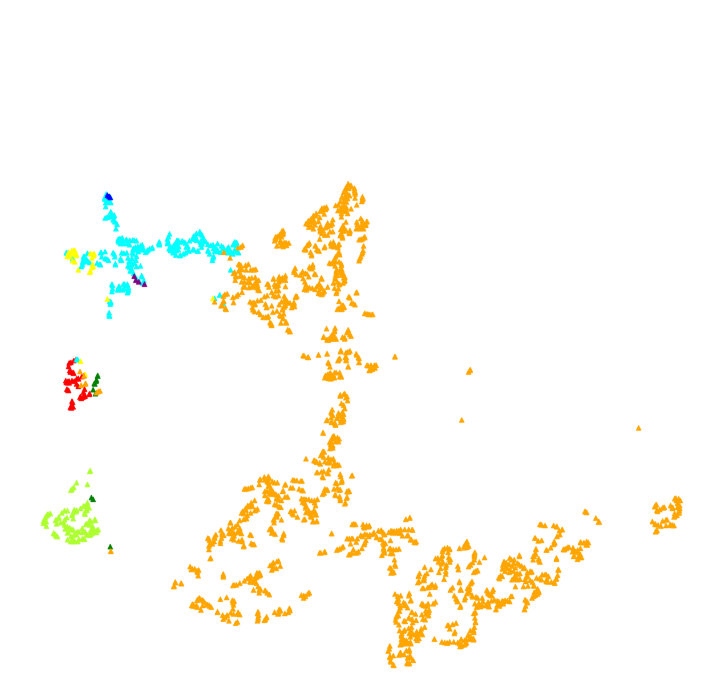}
    \end{minipage}      
     \begin{minipage}  {0.32\linewidth}
        \centering
        \includegraphics [width=0.8\linewidth,height=0.8\linewidth] 
        {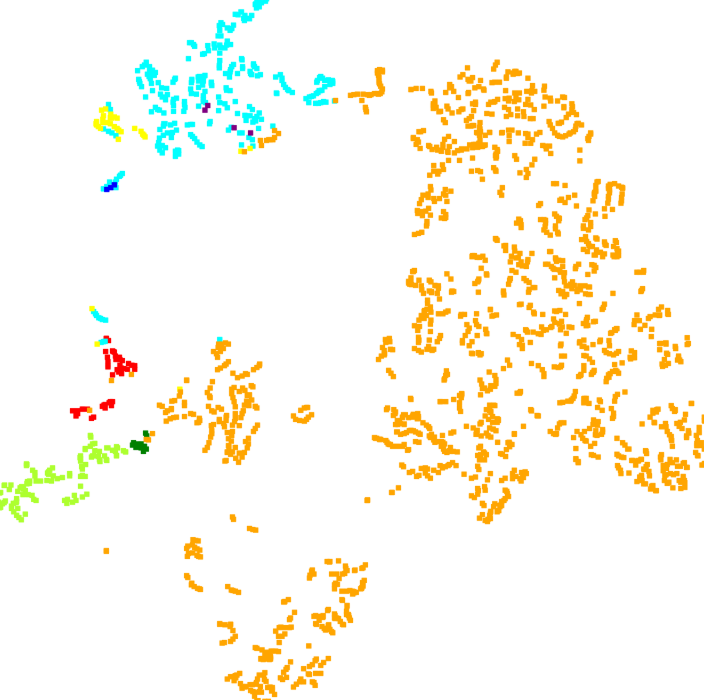}
    \end{minipage} 
	 
	 \vspace{0.1cm}
	 
    \begin{minipage}  {0.32\linewidth}
        \centering
        \includegraphics [width=0.8\linewidth,height=0.8\linewidth] 
        {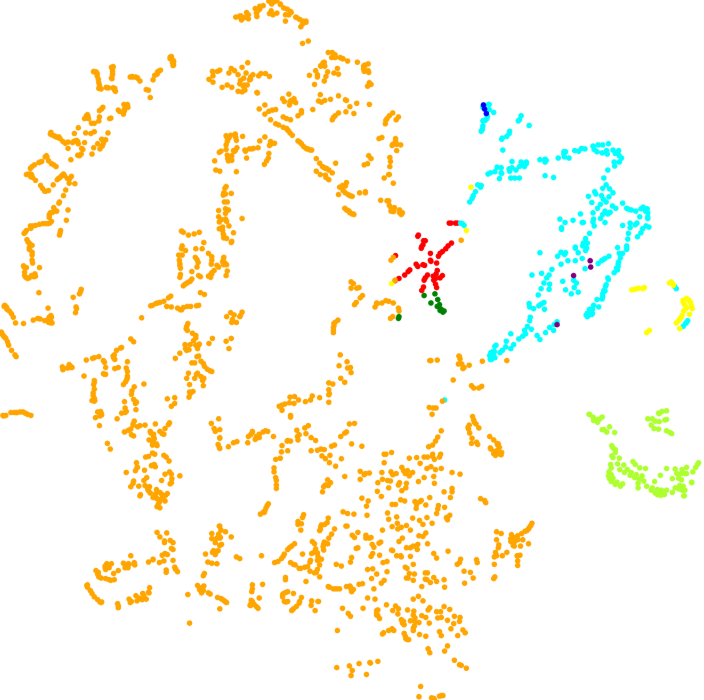}\\
        {\small \uppercase\expandafter{\romannumeral1}. Original}
    \end{minipage}      
    \begin{minipage}  {0.32\linewidth}
        \centering
        \includegraphics [width=0.8\linewidth,height=0.8\linewidth] 
        {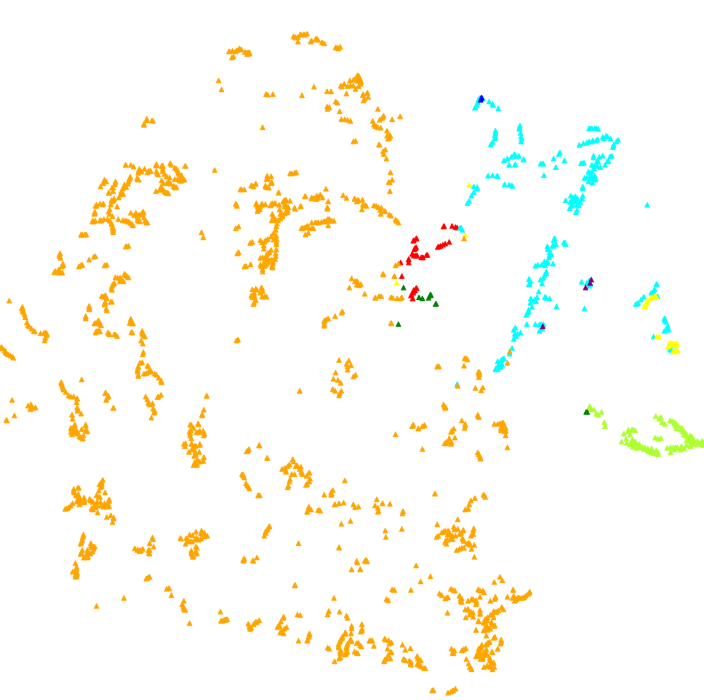}\\
        {\small \uppercase\expandafter{\romannumeral2}. Contextual aug}
    \end{minipage}      
    \begin{minipage}  {0.32\linewidth}
        \centering
        \includegraphics [width=0.8\linewidth,height=0.8\linewidth] 
        {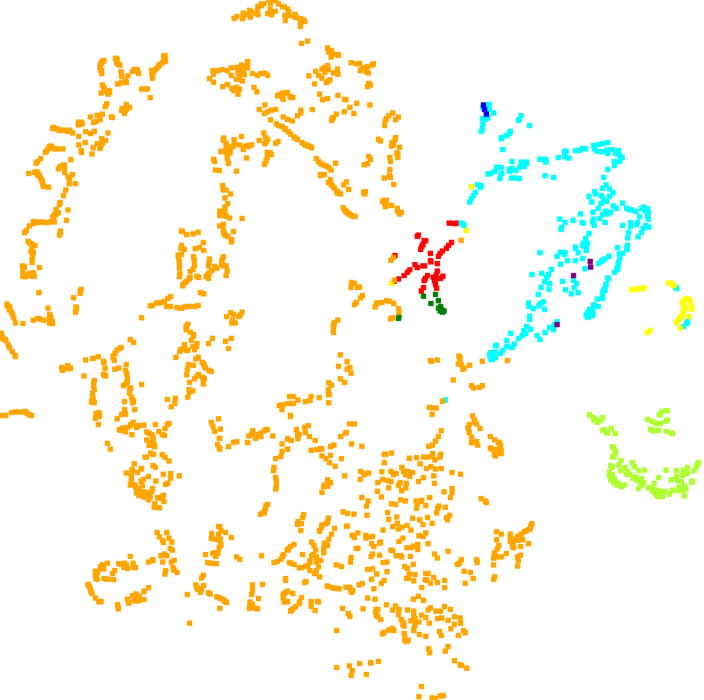}\\
        {\small \uppercase\expandafter{\romannumeral3}. Low-level aug}
    \end{minipage} 
    \vspace{0.2cm}     
          
    \caption{Visual comparison between \textit{contextual augmentation} (\uppercase\expandafter{\romannumeral1} and \uppercase\expandafter{\romannumeral2}) and \textit{low-level augmentation} (\uppercase\expandafter{\romannumeral1} and \uppercase\expandafter{\romannumeral3}) using t-SNE visualization for features of the overlapping region (shown in yellow box).
    \textbf{Top}: input crops from the same image, where \uppercase\expandafter{\romannumeral2} and \uppercase\expandafter{\romannumeral3} apply the \textit{contextual} and \textit{low-level augmentation} respectively.  
    \textbf{Middle}: t-SNE results of the model trained with labeled data only. Note that the three visualizations are in the same t-SNE space, and the dots with the same color represent the features of the same class.
    \textbf{Bottom}: t-SNE results of our method.
    }
    \label{fig:tsne_vis}
\vspace{-0.5cm}
\end{figure}


\begin{figure*}
\begin{center}
\includegraphics[width=0.8\linewidth]{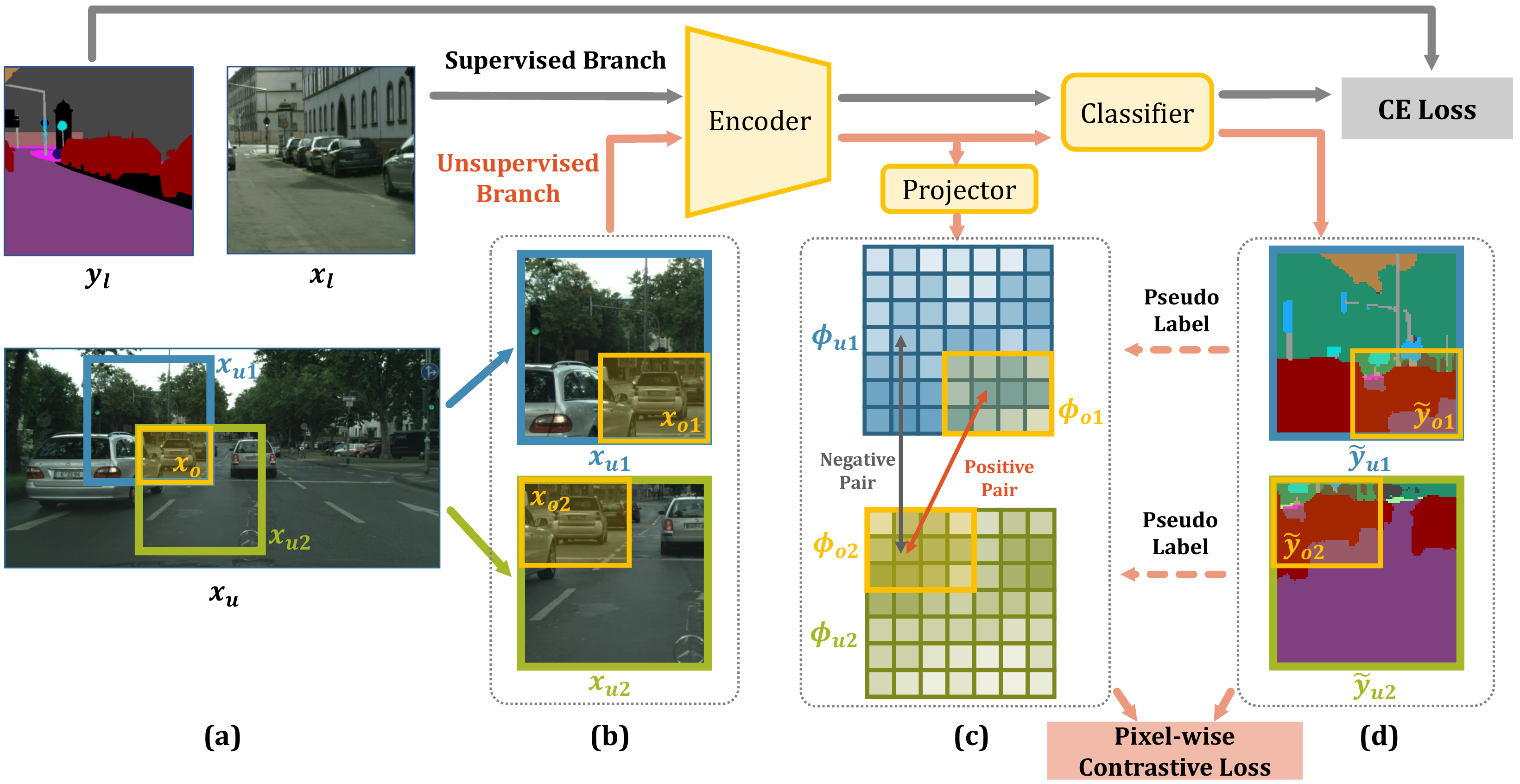}
\end{center}
\vspace{-0.5cm}
\caption{Overview of our framework. In the unsupervised branch, two patches are randomly cropped from the same image with a partially overlapping region. We aim to maintain a pixel-to-pixel consistency between the feature maps corresponding to the overlapping region.}
\label{fig:overview}
\vspace{-0.3cm}
\end{figure*}

In the following sections, we firstly present our motivation in Sec.~\ref{sec:motivation}, and then elaborate the proposed context-aware consistency in Sec.~\ref{sec:consistency}. Also, to accomplish the consistency, we propose the Directional Contrastive Loss in Sec.~\ref{sec:dcloss}. 
Moreover, two sampling strategies further improve the baseline as shown in Sec.~\ref{sec:sampling_startegy}. In Sec.~\ref{sec:weakly-method}, our method generalizes well with extra image-level annotations.

\subsection{Motivation}
\label{sec:motivation}

Consistency-based methods~\cite{temporal_ens,meanteacher,semivat} have achieved decent performance in semi-supervised learning by maintaining the consistency between perturbed images or features to learn robust representations. To accomplish the consistency training in semantic segmentation, one can simply apply low-level data augmentations, such as Gaussian blur and color jitter, to the input images and then constrain the perturbed ones to be consistent. However, low-level augmentations only alter the pixel itself without changing the contextual cues. As shown in the 
\textbf{Middle} row of Fig.~\ref{fig:tsne_vis}, we observe that the embedding distribution changes much more significantly under the \textit{contextual augmentation} (\ie, \uppercase\expandafter{\romannumeral1} and \uppercase\expandafter{\romannumeral2}) than \textit{low-level augmentations} (\ie, \uppercase\expandafter{\romannumeral1} and \uppercase\expandafter{\romannumeral3}). In other words, even when the model has achieved consistency between low-level augmentations, it still could be unable to produce consistent embedding distribution under different contexts, which implies that the consistency with contextual augmentation could be an additional constraint that supplements low-level augmentations. 

Further, one of the reasons why features vary too much under different contexts is that the model overfits the limited training data, causing the features to excessively rely on the contextual cues without sufficient self-awareness.
To this end, maintaining the consistency between features under different contexts can yield more robust features and also help to alleviate the overfitting problem to some extent.

Motivated by the above, we propose our simple yet effective method. Later experiments show that our method surpasses current state-of-the-art methods as well as the method based on only low-level augmentations. 

\subsection{Context-Aware Consistency}
\label{sec:consistency}

The overview of our framework is shown in Fig.~\ref{fig:overview}. Specifically, there are two batches of inputs, \ie, $x_l$ and $x_{u}$, representing labeled and unlabeled data respectively. 
As common semantic segmentation 
models, the labeled images $x_l$ pass through the encoder network $\mathcal E$ to get the feature maps $f_l = \mathcal E(x_l)$. Then, the classifier $\mathcal C$ makes predictions $p_l = \mathcal C(f_l)$, which are supervised by ground truth labels $y_l$ with the standard cross entropy loss $\mathcal L_{ce}$.

As for the unlabeled image $x_{u}$, 
two patches  $x_{u1}$ and $x_{u2}$ are randomly cropped with an overlapping region $x_o$ (Fig.~\ref{fig:overview} (a)).
Then, $x_{u1}$ and $x_{u2}$ are processed by different low-level augmentations, and further pass through the encoder $\mathcal E$ to get the feature maps $f_{u1}$ and $f_{u2}$ respectively (Fig.~\ref{fig:overview} (b)). Next, similar to~\cite{simclr}, they are projected as $\phi_{u1}=\Phi(f_{u1})$ and $\phi_{u2}=\Phi(f_{u2})$ by the non-linear projector $\Phi$. The features of the overlapping region $x_o$ in $\phi_{u1}$ and $\phi_{u2}$ are denoted as $\phi_{o1}$ and $\phi_{o2}$ respectively (Fig.~\ref{fig:overview} (c)). 

The context-aware consistency is then maintained between $\phi_{o1}$ and $\phi_{o2}$ by the Directional Contrastive Loss (DCL) that encourages the representations of the overlapping region $x_o$ to be consistent under different contexts, \ie, the 
non-overlapping regions in $x_{u1}$ and $x_{u2}$.



Even though the context-aware consistency is used, we don't intend to make the features totally ignore the contexts.
Our purpose is just to alleviate the excessive contextual reliance and make the contexts get used more properly.
On the one hand, the supervised loss $\mathcal L_{ce}$ is used to prevent the model from degrading into totally ignoring the contexts. On the other hand, the non-linear projector $\Phi$ is employed for the alignment. 
The projector $\Phi$ projects the features into a lower dimension, and only the projector output features are directly required to be invariant to the contexts rather than the encoder output features. Therefore, the projector actually plays the role of information bottleneck that prevents the original features losing useful contextual information for segmentation. In Table~\ref{tab:ablation}, we also conduct an experiment to highlight the contribution of the projector.
\begin{figure}
\begin{center}
\includegraphics[width=1\linewidth]{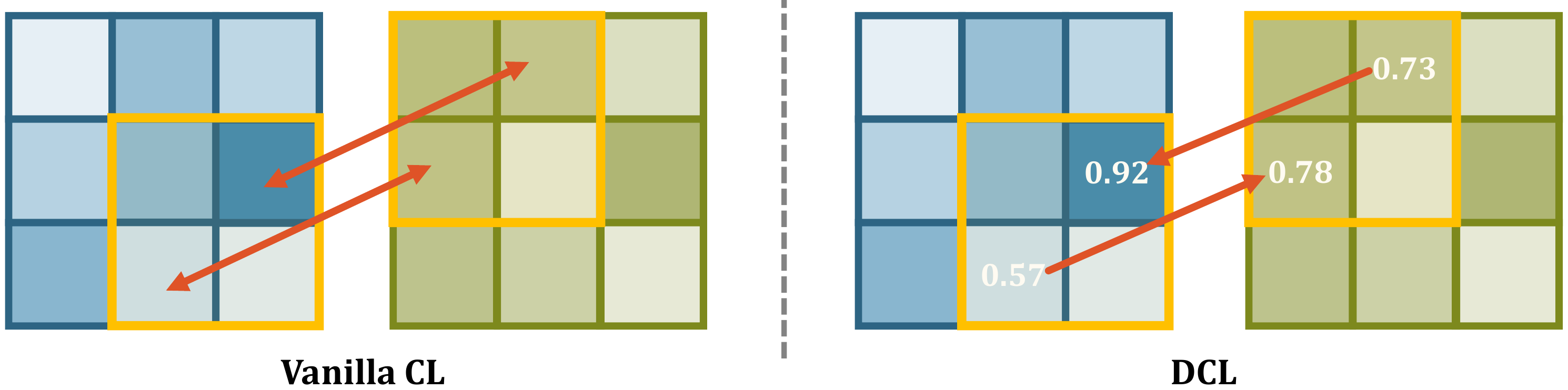}
\end{center}
\vspace{-0.3cm}
\caption{Comparison between vanilla Contrastive Loss (CL) and Directional Contrastive Loss (DCL). Each grid represents a feature. The scalar in the grid means the confidence of that feature.}

\label{fig:dcl}
\vspace{-0.3cm}
\end{figure}

\subsection{Directional Contrastive Loss (DC Loss)}
\label{sec:dcloss}


The context-aware consistency requires each feature in $\phi_{o1}$ to be consistent with the corresponding feature of the same pixel in $\phi_{o2}$, making the high-level features less vulnerable to the varying environments. To accomplish this alignment, the most straightforward solution is to apply $\ell_2$ loss between $\phi_{o1}$ and $\phi_{o2}$. However, $\ell_2$ loss is too weak to make the features discriminative without pushing them away from the negative samples,
which is shown in later experiments (Table~\ref{tab:ablation}). Instead, we take the inspiration from the contrastive loss and propose Directional Contrastive Loss (DC Loss), which accomplishes contrastive learning in the pixel level.


Unlike the $\ell_2$ loss, DC Loss not only forces the positive samples, \ie, the features with the same class, to lie closer, but also separates the negative samples that belong to other classes. Specifically, as shown in Fig.~\ref{fig:overview} (c), 
we regard two features at the same location of $\phi_{o1}$ and $\phi_{o2}$ as a positive pair, because they both correspond to the same pixels in $x_{u}$ but under different contexts, \ie, the 
non-overlapping regions in $x_{u1}$ and $x_{u2}$. 
In addition, any two features in $\phi_{u1}$ and $\phi_{u2}$ at different locations of the original image can be deemed as a negative pair (in Fig.~\ref{fig:overview} (c)).


Furthermore, the proposed DC Loss additionally incorporates a directional alignment for the positive pairs.
Specifically, we compute the maximum probability among all classes, \ie, $\max(\mathcal{C}(f_i))$, as the confidence of each feature $\phi_i$, where $\mathcal C$ is the classifier. As the prediction with higher confidence generally is more accurate as shown in the supplementary file, the less confident feature is required to be aligned towards the more confident counterpart (in Fig.~\ref{fig:dcl}), which effectively prevents the more confident feature from corrupting towards the less confident one.

Formally, for the $b$-th unlabeled image, the DC Loss $\mathcal L_{dc}^b$ can be written as follows.

\vspace{-0.4cm}

\begin{footnotesize}
\begin{align}
\label{loss:vanilla_dcloss}
 & l_{dc}^{b}(\phi_{o1}, \phi_{o2}) = \notag \\
 & -\frac{1}{N}\sum_{h,w} \mathcal{M}^{h, w}_{d} \cdot \log \frac{r(\phi_{o1}^{h,w}, \phi_{o2}^{h,w}) }{r(\phi_{o1}^{h,w}, \phi_{o2}^{h,w}) + \sum\limits_{\phi_{n} \in \mathcal{F}_u}{r(\phi_{o1}^{h,w}, \phi_{n})}}
\end{align}

\vspace{-0.3cm}

\begin{align}
\label{loss:dir_mask}
\mathcal{M}^{h, w}_{d} &= \mathbf{1}\{\max\mathcal{C}(f_{o1}^{h,w}) < \max\mathcal{C}(f_{o2}^{h,w})\} \\
\mathcal{L}_{dc}^{b} &= l_{dc}^{b}(\phi_{o1}, \phi_{o2}) + l_{dc}^{b}(\phi_{o2}, \phi_{o1})
\end{align}
\end{footnotesize}


\vspace{-0.5cm}

where $r$ denotes the exponential function of the cosine similarity $s$ between two features with a temperature $\tau$, \ie, $ r(\phi_1, \phi_2) = \exp{(s(\phi_1, \phi_2) / \tau)}$, $h$ and $w$ denote the 2-D spatial locations, $N$ denotes the number of spatial locations of the overlapping region, $\phi_{n} \in \mathbb{R}^{c}$ represents the negative counterpart of the feature $\phi_{o1}^{h,w}$, and $\mathcal{F}_u$ represents the set of negative samples.
We note that the gradients of $l_{dc}^{b}(\phi_{o1}, \phi_{o2})$ are only back propagated to $\phi_{o1}^{h,w}$.
As we observe that more negative samples lead to better performance, we select the negative samples $\phi_{n}$ not only from the current image, but also from all unlabeled images within the current training batch.
Moreover, to further increase negative samples, we maintain a memory bank to store the features in the past few batches to get sufficient negative samples. We emphasize that the increasing negative samples only incurs minor additional computation and memory cost with our implementation, which is demonstrated in later experiments in Sec.~\ref{exp:ablation_ns}.

\vspace{-0.3cm}
\paragraph{Comparison with Vanilla Contrastive Loss} The proposed Directional Contrastive Loss (DCL) differs from vanilla Contrastive Loss (CL) mainly in two ways. Firstly, CL is applied to the image-level feature, while DCL conducts contrastive learning in a pixel-wise manner.
Secondly, CL as well as the Supervised Contrastive Loss~\cite{sup_contrast} does not consider the confidence of features, and simply aligns them with each other bilaterally, which may even corrupt the better feature by forcing it to align towards the worse one. However, DCL only requires the less confident feature to be aligned towards the more confident counterpart.


\subsection{Sampling Strategies}
\label{sec:sampling_startegy}

\begin{figure}
	\centering
    \begin{minipage}  {0.7\linewidth}
        \centering
        \includegraphics [width=1\linewidth,height=0.5\linewidth] 
        {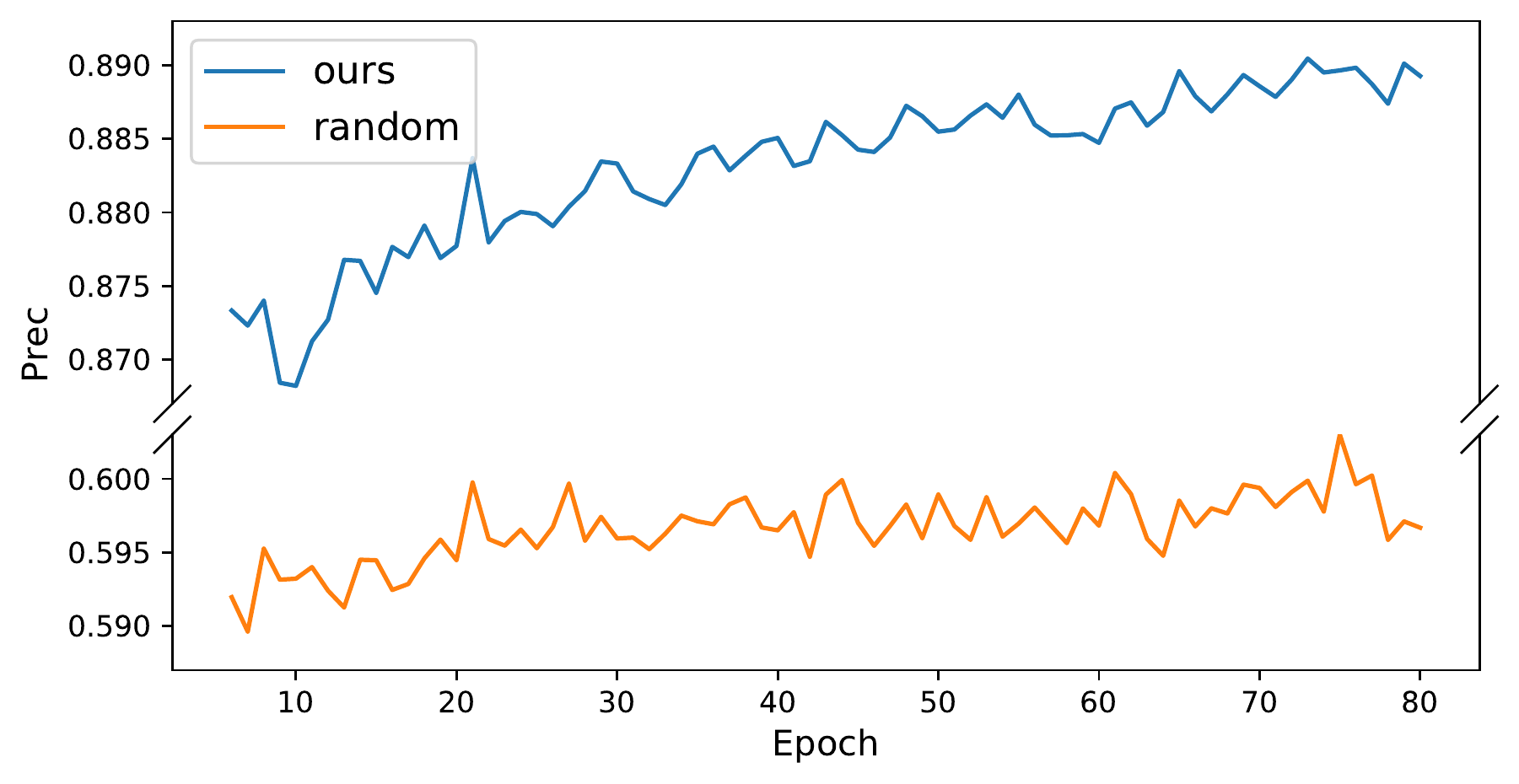}
    \end{minipage}
	 \vspace{-0.1cm}
    \caption{The precision of selected negative samples in each training epoch. The orange curve represents the result of random sampling, while the blue one represents the result of our negative sampling strategy. Note that the precision is computed by dividing the number of true negative samples by the number of all selected negative samples. Best viewed in zoom.}
    \label{fig:prec}
\vspace{-0.3cm}
\end{figure}


\paragraph{Negative Sampling} 
Although image-level contrastive learning has made significant progress, it is hard to transfer the image-level success to pixel-level tasks like semantic segmentation. Because in segmentation, many different pixels in an image may belong to the same class, especially in the case of the background class or large objects such as sky and sidewalk. Therefore, when randomly selecting negative samples for an anchor feature, it is very common to select false negative samples, \ie, those features that actually belong to the same class as the anchor feature. As shown in the orange curve of Fig.~\ref{fig:prec}, random sampling causes that only less than 60\% of the negative samples are actually true. 
In this case, the false negative pairs are forced to be separated from each other, which adversely affects or even corrupts the representation learning. 


To avoid the common false negative pairs, we make use of pseudo labels as heuristics to eliminate those negative samples with high probability to be false. Specifically, when selecting negative samples, we also compute the probability for each class by forwarding feature maps of unlabeled data, \ie, $f_{u1}$ and $f_{u2}$, into the classifier $\mathcal C$, then get the class indexes with the highest probability, \ie, $\tilde{y}_{u1}$ and $\tilde{y}_{u2}$ (Fig. ~\ref{fig:overview} (d)). Formally, we have

\vspace{-0.1cm}
\begin{footnotesize}
\begin{equation}
    \tilde{y}_{ui} = \mathop{\arg\max} \  \mathcal C(f_{ui}) \qquad i \in \{1,2\}
\end{equation}
\end{footnotesize}

\vspace{-0.3cm}
For an anchor feature $\phi_{o}^{h,w}$ with pseudo label $\tilde{y}_{o}^{h,w}$, the selected negative samples should have different pseudo labels ($\tilde{y}_{n} \ne \tilde{y}_{o}^{h,w}$).  Hence, the original equation (2) is accordingly updated as
\begin{footnotesize}
\begin{align}
    \label{eqn:neg_sample}
    &l_{dc}^{b, ns}(\phi_{o1}, \phi_{o2}) = 
    \notag \\
    & -\frac{1}{N} \sum_{h,w} \mathcal{M}^{h,w}_{d} \cdot  \log
     \frac{r(\phi_{o1}^{h,w}, \phi_{o2}^{h,w})}{r(\phi_{o1}^{h,w}, \phi_{o2}^{h,w}) + \sum\limits_{\phi_n \in \mathcal{F}_u}{\mathcal M_{n,1}^{h,w} \cdot r(\phi_{o1}^{h,w}, \phi_{n})}}
\end{align}
\end{footnotesize}
\vspace{-0.3cm}

where $\mathcal M_{n,1}^{h,w} = \mathbf{1}\{\tilde{y}_{o1}^{h,w} \ne \tilde{y}_{n} \}$ is a binary mask indicating whether the pseudo labels $\tilde{y}_{o1}^{h,w}$ and $\tilde{y}_{n}$ for the two features $\phi_{o1}^{h,w}$ and $\phi_n$ are different.

As shown in Fig.~\ref{fig:prec}, by exploiting pseudo labels to eliminate the false negative samples, the precision increased from around 60\% to 89\%. As most of the false negative pairs are filtered out, the training process becomes more stable and robust. Experiments in Table~\ref{tab:ablation} show the huge improvement.
 

\vspace{-0.3cm}
\paragraph{Positive Filtering}
Although we enable the less confident feature to align towards the more confident counterpart in the DC Loss, it may still cause the less confident feature to corrupt if the more confident counterpart is not confident enough. Therefore, to avoid this case, we also filter out those positive samples with low confidence. In particular, if the confidence of a positive sample is lower than a threshold $\gamma$, then this positive pair will not contribute to the final loss. Formally, the Eq. \eqref{eqn:neg_sample} is further revised as 

\vspace{-0.3cm}
\begin{footnotesize}
\begin{align}
    \label{eqn:pos_filter}
    & l_{dc}^{b, ns, pf}(\phi_{o1}, \phi_{o2}) = \notag \\ & -\frac{1}{N} \sum_{h,w} \mathcal{M}^{h,w}_{d,pf} \cdot \log \frac{r(\phi_{o1}^{h,w}, \phi_{o2}^{h,w})}{r(\phi_{o1}^{h,w}, \phi_{o2}^{h,w}) + \sum\limits_{\phi_n \in \mathcal{F}_u}{ \mathcal M_{n,1}^{h,w} \cdot r(\phi_{o1}^{h,w}, \phi_{n})}}
\end{align}

\vspace{-0.3cm}

\begin{equation}
    \mathcal{M}^{h,w}_{d,pf} = \mathcal{M}^{h,w}_{d} \cdot \mathbf{1}\{\max \mathcal C(f_{o2}^{h,w}) > \gamma\}
\end{equation}
\end{footnotesize}
where $\mathcal{M}^{h,w}_{d,pf}$ is the binary mask that not only considers the directional mask $\mathcal{M}^{h,w}_{d}$, but also filters those uncertain positive samples.
So, we have our final loss
\vspace{-0.2cm}
\begin{footnotesize}
\begin{equation}
	\mathcal L_{dc}^{ns, pf} = \frac{1}{B} \sum\limits_{b=1}^{B}( l_{dc}^{b, ns, pf}(\phi_{o1}, \phi_{o2}) + l_{dc}^{b, ns, pf}(\phi_{o2}, \phi_{o1}))
\end{equation}
\vspace{-0.3cm}
\begin{equation}
	\mathcal L = \mathcal L_{ce} + \lambda \mathcal L_{dc}^{ns, pf}
\end{equation}
\end{footnotesize}
where $B$ represents the training batch size, $\lambda$ controls the contribution of the unsupervised loss.


\subsection{Extension with Extra Image-level Annotations}
\label{sec:weakly-method}
Practically, common weak annotations such as image-level labels, bounding boxes and scribbles can be exploited to further boost the performance. 
Our method can be easily adapted to the setting where a small set of pixel-level labeled data and a much larger collection of image-level labeled data are provided.

Following the previous work~\cite{CCT}, 
we first pre-train a classification network with the weakly labeled data, and then the pseudo labels $y_p$ can be acquired by CAM~\cite{cam}. During training, in addition to the main classifier $\mathcal C$ which is used to make predictions for the pixel-level labeled data, we add an extra classifier $\mathcal C_w$ for the weakly annotated images whose pseudo labels $y_p$ are relatively coarse and inaccurate. In this way, it prevents the coarse labels from corrupting the main classifier $\mathcal C$. We keep all other components the same as semi-supervised setting, so the loss function becomes

\begin{footnotesize}
\begin{equation}
    \mathcal L = \mathcal L_{ce} + \lambda \mathcal L_{dc}^{ns, pf} + \lambda_w \mathcal L_{w}   
\end{equation}
\vspace{-0.2cm}
\begin{equation}
    \mathcal L_{w} = \frac{1}{2} \cdot( CE(\mathcal C_w(f_{u1}), y_p) + CE(\mathcal C_w(f_{u2}), y_p))
\end{equation}
\end{footnotesize}
 where $\mathcal L_{w}$ is the weakly-supervised loss and we follow CCT~\cite{CCT} to set up the weighting factor $\lambda_w$. During inference, we simply discard the auxiliary classifier $\mathcal C_w$.




\section{Experiments}
In the following, we show the implementation details in Sec.~\ref{sec:setup}, followed by the comparison with state-of-the-art methods in Sec.~\ref{sec:comparison}. The ablation study is presented in Sec.~\ref{sec:ablation} and we further apply our method with extra image-level labels in Sec.~\ref{sec:weakly} to show the adaptation ability.

\subsection{Implementation Details} 
\label{sec:setup}
\paragraph{Network Architecture} Some previous works~\cite{AdvSemiSeg, gct, s4gan_mittal} use DeepLabv2~\cite{deeplabv2} as their base models, while some other methods are verified with different base models, \eg, \cite{errorcorrect} and CCT \cite{CCT} use DeepLabv3+~\cite{deeplabv3+} and PSPNet~\cite{pspnet} respectively. Since DeepLabv3+ shares a very similar structure with DeepLabv2 and the former achieves better performance, we use DeepLabv3+ as the main segmentation network. We also implement our method on PSPNet~\cite{pspnet} to show the generalization ability.

The encoder in Fig.~\ref{fig:overview} refers to all other components except the final classifier, \ie, the encoders of PSPNet and DeepLabv3+ also include the PPM and ASPP modules respectively.
The projector $\Phi$ is implemented by an MLP that consists of two FC layers (128 output channels) and one intermediate ReLU layer. 

Different segmentation models vary in their output resolutions. For DeepLabv3+, the spatial size of the output feature map before the classifier is 1/4 of the input image. To be more memory-efficient, the feature map is firstly down-sampled via an $2\times 2$ average pooling layer and then sent to the projector $\Phi$. For PSPNet, the spatial size of output feature map is 1/8 of input images, so the feature map is directly sent to $\Phi$ without additional processing.

\vspace{-0.4cm}
\paragraph{Datasets} The experiments are conducted on PASCAL VOC~\cite{pascalvoc,pascalvocaug} and Cityscapes~\cite{cityscapes}. The details are illustrated in the supplementary file.

 

\vspace{-0.4cm}
\paragraph{Experimental Setting}
During training, labeled images are first randomly resized by a ratio between 0.5 and 2 followed by the random cropping ($320 \times 320$ for PASCAL VOC and $720 \times 720$ for Cityscapes). Then we apply the horizontal flip with a probability of 0.5 to the cropped patches. Differently, for each unlabeled image, after being randomly resized, two patches $x_{u1}$ and $x_{u2}$ are randomly cropped from the same resized image and the Intersection-over-Union(IoU) value of these two patches is supposed to be within the range  $[0.1, 1.0]$. 
After that, we apply random mirror and standard pixel-wise augmentations (\ie, Gaussian blur, color jitters and gray scaling) to $x_{u1}$ and $x_{u2}$. 

Following the common practice, we use `poly' learning rate decay policy where the base learning rate is scaled by $(1-iter/max\_iter)^{power}$ and $power$ is set to 0.9 in our experiments. SGD optimizer is implemented with weight decay 0.0001. The base learning rate values are set to 0.001 and 0.01 for backbone parameters and the others respectively for PASCAL VOC, while 0.01 and 0.1 for Cityscapes. The temperature $\tau$ is set to 0.1. Two NVIDIA GeForce RTX 2080Ti GPUs are used for training our method unless specified for PASCAL VOC while four are used for Cityscapes, and a training batch includes 8 labeled and 8 unlabeled images. The unsupervised loss weight $\lambda$ and the threshold for positive filtering $\gamma$ are set to 0.1 and 0.75. To stabilize training, we only use the supervised cross entropy loss to train the main segmentation model in the first 5 epochs. All models are trained entirely for 80 epochs on both datasets. 

The mean Intersection-over-Union (mIoU) is adopted as our evaluation metric and each image is tested in its original size. We compare our methods in the settings with different labeled data proportions, \ie, full, 1/4, 1/8 and 1/16, and all results are averaged over 3 runs. 
Note that in the full data setting, images fed to the unsupervised branch are simply collected from the labeled set.

\begin{table}
\tabcolsep=0.15cm
\begin{center}
\begin{footnotesize}
\begin{tabular}{l l l c c c c }
\hline
\toprule Method & SegNet & Backbone & 1/16 & 1/8 & 1/4 & Full \\
\specialrule{0em}{2pt}{0pt}
\hline
\specialrule{0em}{2pt}{0pt}
SupOnly & PSPNet & ResNet50 & 57.4 & 65.0 & 68.3 & 75.1 \\
CCT~\cite{CCT} & PSPNet & ResNet50 & 62.2 & 68.8 & 71.2 & 75.3 \\
Ours & PSPNet & ResNet50 & \textbf{67.1} & \textbf{71.3} & \textbf{72.5} & \textbf{76.4} \\
\specialrule{0em}{2pt}{0pt}
\hline
\specialrule{0em}{2pt}{0pt}
SupOnly & DeepLabv3+ & ResNet50 & 63.9 & 68.3 & 71.2 & 76.3 \\
ECS~\cite{errorcorrect} & DeepLabv3+ & ResNet50 & - & 70.2 & 72.6 & 76.3\\
Ours & DeepLabv3+ & ResNet50 & \textbf{70.1} & \textbf{72.4} & \textbf{74.0} & \textbf{76.5} \\
\specialrule{0em}{2pt}{0pt}
\hline
\specialrule{0em}{2pt}{0pt}
SupOnly & DeepLabv3+ & ResNet101 & 66.4 & 71.0 & 73.5 & 77.7 \\
S4GAN~\cite{s4gan_mittal} & DeepLabv3+ & ResNet101 & 69.1 & 72.4 & 74.5 & 77.3  \\
GCT~\cite{gct} & DeepLabv3+ & ResNet101 & 67.2 & 72.5 & 75.1 & 77.5 \\
Ours & DeepLabv3+ & ResNet101 & \textbf{72.4} & \textbf{74.6} & \textbf{76.3} & \textbf{78.2} \\
\bottomrule                         

\end{tabular}
\end{footnotesize}
\end{center}
\vspace{-0.2cm}
\caption{Comparison with the baseline (SupOnly, \ie, with only supervised loss) and current state-of-the-art methods evaluated on PASCAL VOC with 1/16, 1/8, 1/4 and full labeled data. We use DeepLabv3+ and PSPNet as the main segmentation network, ResNet101 and ResNet50~\cite{resnet} as the backbone.}
\label{table:comparison_voc}
\vspace{-0.3cm}
\end{table}


\begin{table}
    \vspace{0.3cm}
    \centering
    \tabcolsep=0.3cm
    {
        \begin{footnotesize}
        \begin{tabular}{ 
           l
           l
           l
           c  }
            \toprule
            Methods
            & 1/8
            & 1/4
            & Full  \\

            \specialrule{0em}{0pt}{2pt}
            \hline
            \specialrule{0em}{2pt}{0pt}
            
            SupOnly & 66.0 & 70.7 & \textbf{77.7} \\ 
            
            Ours & \textbf{69.7} & \textbf{72.7} & 77.5 \\
            
            \bottomrule                                   
        \end{tabular}
        \end{footnotesize}
    }
    \vspace{0.1cm}
    \caption{Comparison with SupOnly on Cityscapes with 1/8, 1/4 and full labeled data. All results are based on Deeplabv3+~\cite{deeplabv3+} with ResNet-50 backbone.} 
    \label{table:comparison_city}
\vspace{-0.3cm} 
\end{table}

\subsection{Results}
\label{sec:comparison}
To demonstrate the superiority of our method, we make comparisons with recent state-of-the-art models. However, it is hard to compare these methods that are implemented with various settings, \eg, different segmentation models, randomly sampled data lists and inconsistent baseline performance. Therefore, we reproduce the representative models ~\cite{s4gan_mittal, gct, CCT} within an unified framework according to their official code, where all methods are applied upon the same base segmentation model and trained with the same data lists. As the implementation of ECS~\cite{errorcorrect} is not publicly available, we directly use the results reported in the original paper. It is worth noting that our reproduced results are better than those reported in the original papers, and we will also release our code publicly. 

The comparison on PASCAL VOC is shown in Table~\ref{table:comparison_voc}, where our model surpasses other methods by a large margin. 
S4GAN~\cite{s4gan_mittal} uses an additional discriminator to obtain extra supervision, and both GCT~\cite{gct} and ECS~\cite{errorcorrect} refine the flaw or error by exploiting unlabeled images. However, they do not explicitly maintain the contextual consistency with the unlabeled images. Although CCT~\cite{CCT} enables features with different contexts to be consistent by aligning the perturbed high-level features to the main features, the perturbation directly applied to features is unnatural, and also the alignment does not push away the features in different classes. Moreover, in Table~\ref{table:comparison_city}, the experimental results on Cityscapes further demonstrate the generalization ability of our method.

\begin{figure}
	\centering
    \begin{minipage}  {0.48\linewidth}
        \centering
        \includegraphics [width=1\linewidth,height=0.74\linewidth] 
        {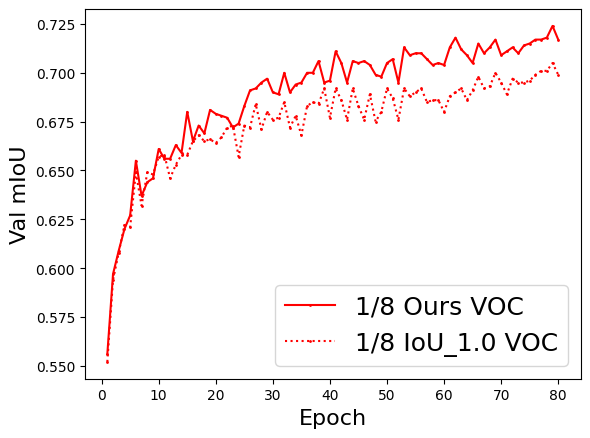}
    \end{minipage}      
    \begin{minipage}  {0.48\linewidth}
        \centering
        \includegraphics [width=1\linewidth,height=0.74\linewidth] 
        {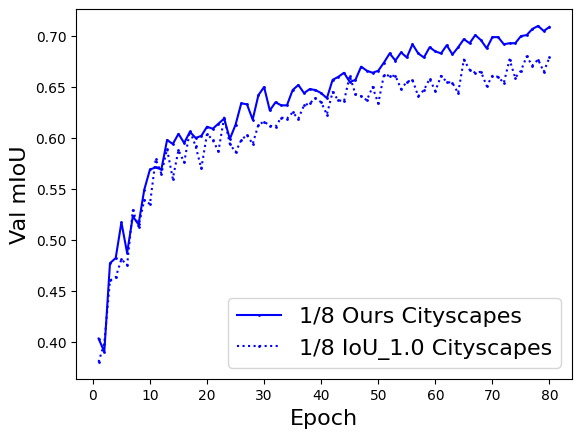}
    \end{minipage}
	 
    \caption{Performance evaluated on the validation sets of PASCAL VOC and Cityscapes respectively during training. IoU\_1.0 means that two patches of unlabeled data totally overlap and only low-level augmentations are applied.}
    \label{fig:contextual_vis}
\end{figure}

\begin{figure*}
	\centering
    \begin{minipage}  {0.13\linewidth}
        \centering
        \includegraphics [width=1\linewidth,height=0.75\linewidth]
        {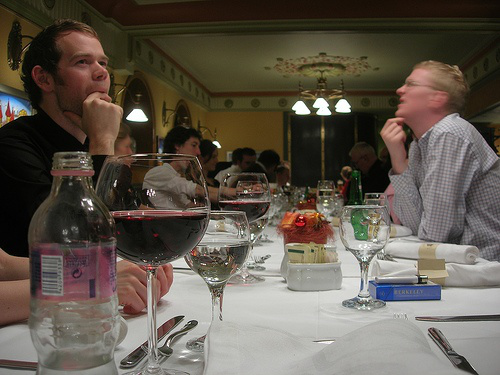}
    \end{minipage}      
    \begin{minipage}  {0.13\linewidth}
        \centering
        \includegraphics [width=1\linewidth,height=0.75\linewidth]
        {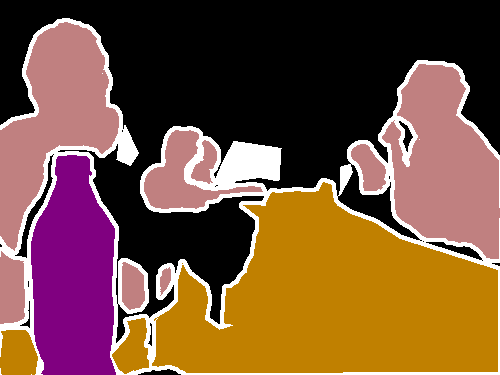}
    \end{minipage}      
     \begin{minipage}  {0.13\linewidth}
        \centering
        \includegraphics [width=1\linewidth,height=0.75\linewidth]
        {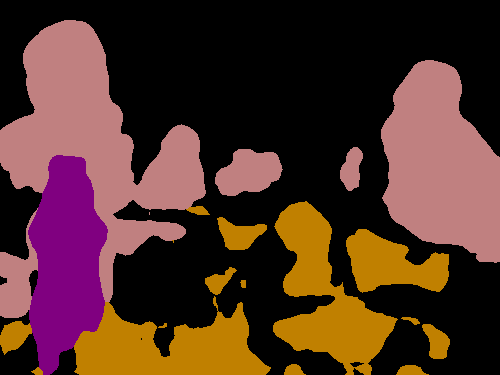}
    \end{minipage} 
     \begin{minipage}  {0.13\linewidth}
        \centering
        \includegraphics [width=1\linewidth,height=0.75\linewidth]
        {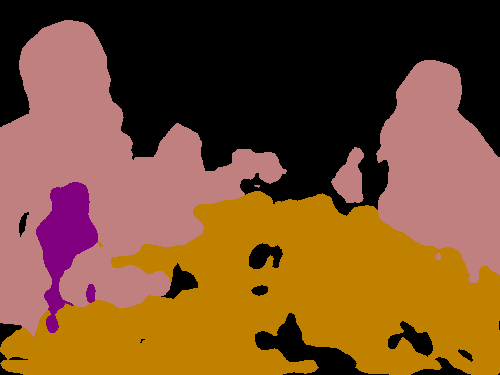}
    \end{minipage} 
     \begin{minipage}  {0.13\linewidth}
        \centering
        \includegraphics [width=1\linewidth,height=0.75\linewidth]
        {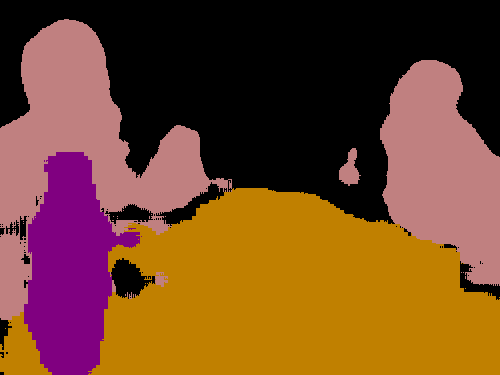}
    \end{minipage} 
     \begin{minipage}  {0.13\linewidth}
        \centering
        \includegraphics [width=1\linewidth,height=0.75\linewidth]
        {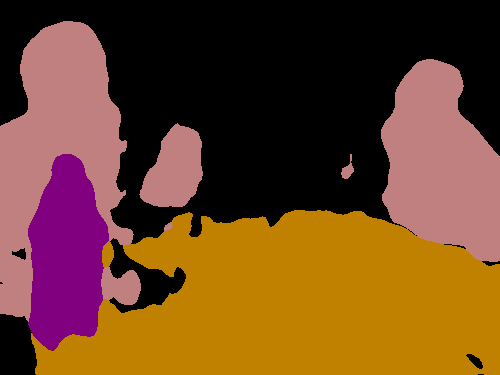}
    \end{minipage} 
     \begin{minipage}  {0.13\linewidth}
        \centering
        \includegraphics [width=1\linewidth,height=0.75\linewidth]
        {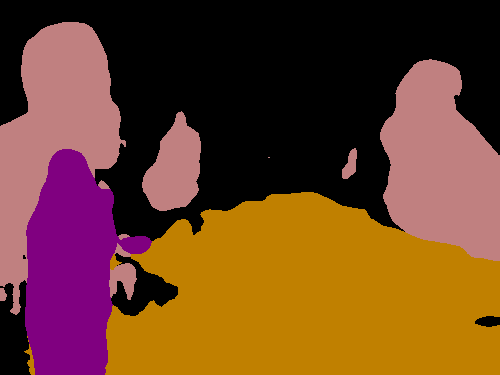}
    \end{minipage} 
	 
	 
    \begin{minipage}  {0.13\linewidth}
        \centering
        \includegraphics [width=1\linewidth,height=0.75\linewidth]
        {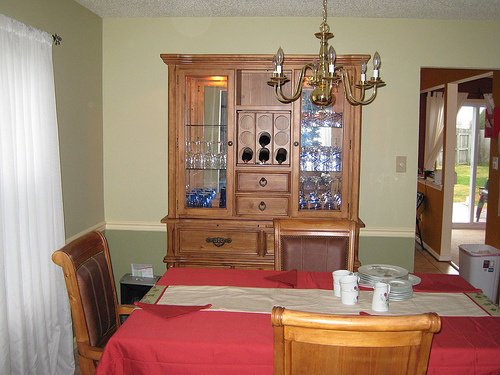}
    \end{minipage}      
    \begin{minipage}  {0.13\linewidth}
        \centering
        \includegraphics [width=1\linewidth,height=0.75\linewidth]
        {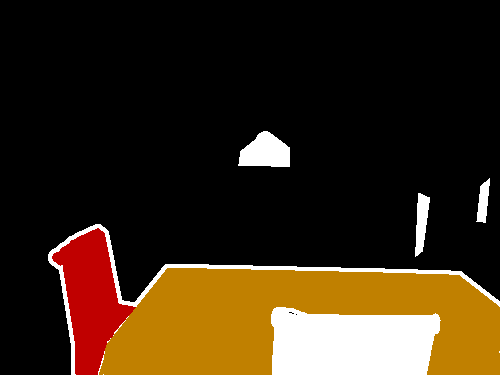}
    \end{minipage}      
     \begin{minipage}  {0.13\linewidth}
        \centering
        \includegraphics [width=1\linewidth,height=0.75\linewidth]
        {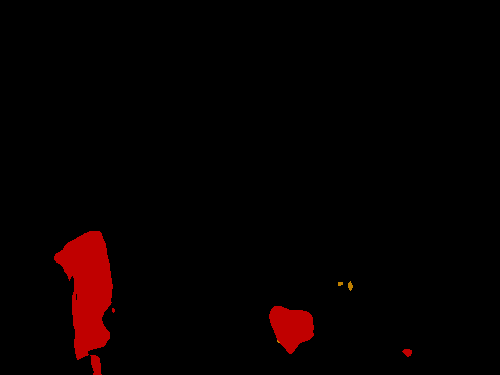}
    \end{minipage} 
     \begin{minipage}  {0.13\linewidth}
        \centering
        \includegraphics [width=1\linewidth,height=0.75\linewidth]
        {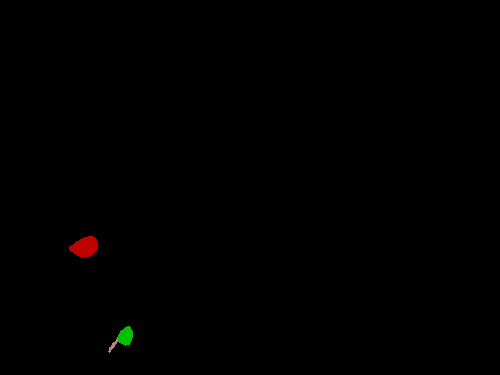}
    \end{minipage} 
     \begin{minipage}  {0.13\linewidth}
        \centering
        \includegraphics [width=1\linewidth,height=0.75\linewidth]
        {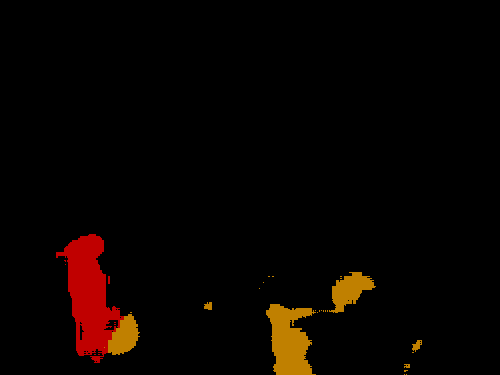}
    \end{minipage} 
     \begin{minipage}  {0.13\linewidth}
        \centering
        \includegraphics [width=1\linewidth,height=0.75\linewidth]
        {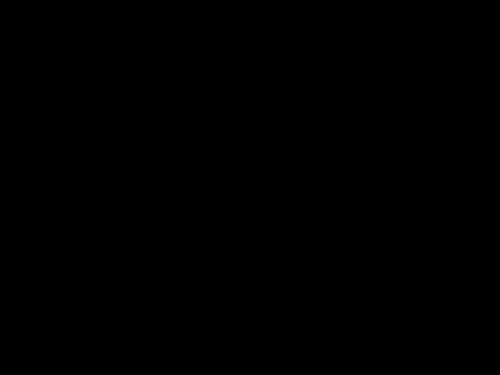}
    \end{minipage} 
     \begin{minipage}  {0.13\linewidth}
        \centering
        \includegraphics [width=1\linewidth,height=0.75\linewidth]
        {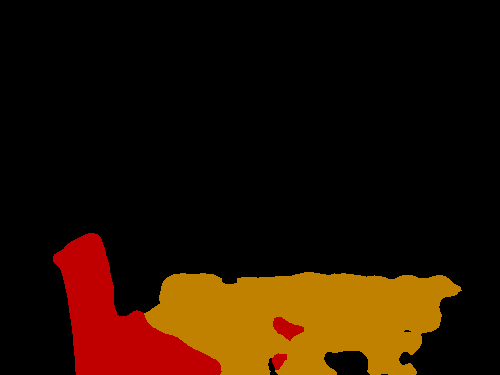}
    \end{minipage} 
	 
	 
    \begin{minipage}[t]{0.13\linewidth}
        \centering
        \includegraphics [width=1\linewidth,height=0.75\linewidth]
        {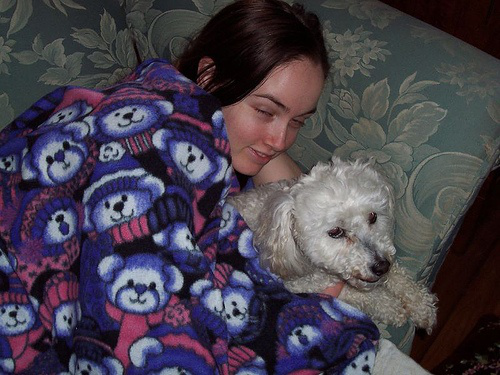}\\\footnotesize{Input Image}
    \end{minipage}      
    \begin{minipage}[t]{0.13\linewidth}
        \centering
        \includegraphics [width=1\linewidth,height=0.75\linewidth]
        {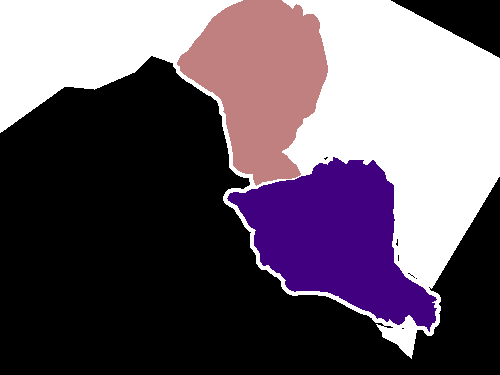}\\\footnotesize{Ground Truth}
    \end{minipage}      
     \begin{minipage}[t]{0.13\linewidth}
        \centering
        \includegraphics [width=1\linewidth,height=0.75\linewidth]
        {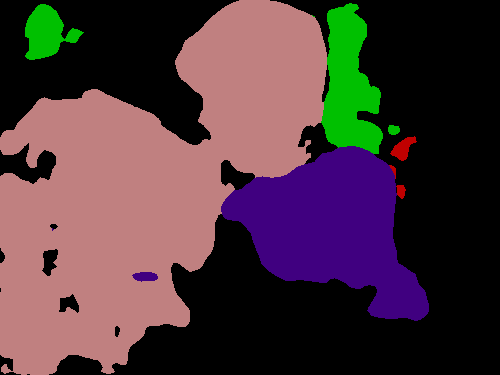}\\\footnotesize{SupOnly}
    \end{minipage} 
     \begin{minipage}[t]{0.13\linewidth}
        \centering
        \includegraphics [width=1\linewidth,height=0.75\linewidth]
        {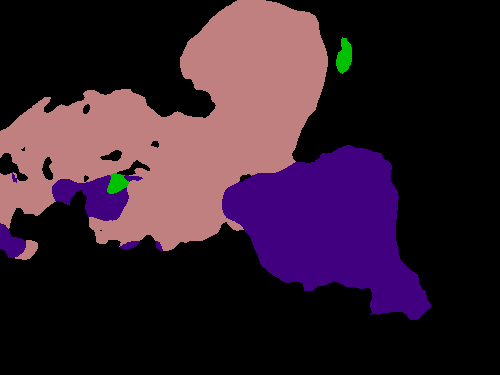}\\\footnotesize{S4GAN}
    \end{minipage} 
     \begin{minipage}[t]{0.13\linewidth}
        \centering
        \includegraphics [width=1\linewidth,height=0.75\linewidth]
        {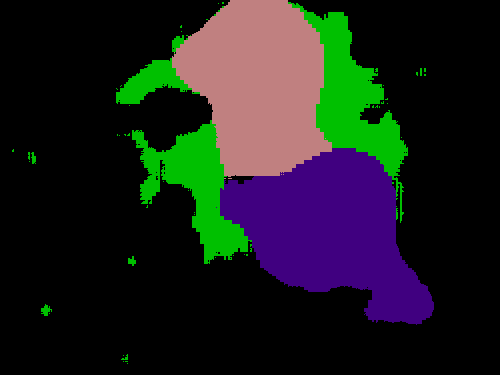}\\\footnotesize{CCT}
    \end{minipage} 
     \begin{minipage}[t]{0.13\linewidth}
        \centering
        \includegraphics [width=1\linewidth,height=0.75\linewidth]
        {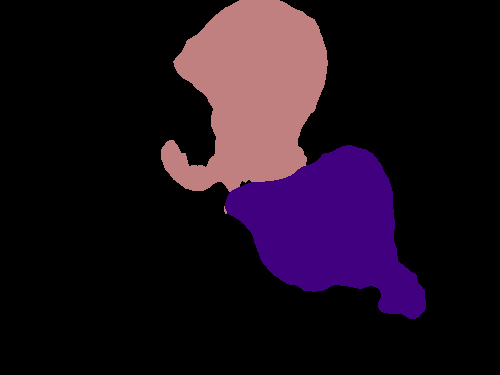}\\\footnotesize{GCT}
    \end{minipage} 
     \begin{minipage}[t]{0.13\linewidth}
        \centering
        \includegraphics [width=1\linewidth,height=0.75\linewidth]
        {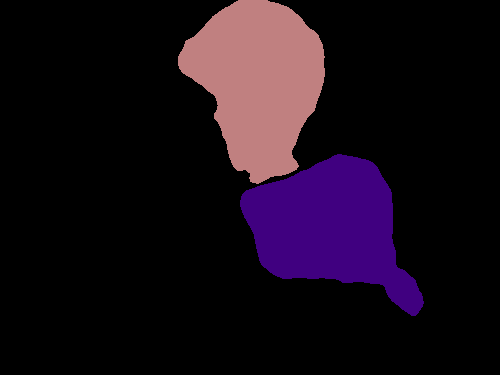}\\\footnotesize{Ours}
    \end{minipage} 
    
    \vspace{0.1cm}     
    \caption{Visual comparison between SupOnly (\ie, trained with only supervised loss) and current state-of-the-art methods with ours.}
    \label{fig:vis_comparison}
\vspace{-0.3cm}
\end{figure*}

\subsection{Ablation Study}
\label{sec:ablation}

\begin{table}[!t]
    \centering
    \tabcolsep=0.15cm
    {
        \begin{footnotesize}
        \begin{tabular}{ c | c  c  c  c  c  c  | c}
            \toprule
            ID & Proj & Context & CL & Dir & NS & PF & mIoU \\

            \specialrule{0em}{0pt}{1pt}
            \hline
            \specialrule{0em}{0pt}{1pt}
            
            SupOnly & & & & & & & 64.7 \\ 
            ST & & & & & & & 66.3 \\ 
            
            \specialrule{0em}{0pt}{1pt}
            \hline
            \specialrule{0em}{1pt}{0pt}
            
            \uppercase\expandafter{\romannumeral1} & \Checkmark & \Checkmark & & & & & 64.2\\ 
            
            \uppercase\expandafter{\romannumeral2} & \Checkmark & \Checkmark & \Checkmark & & & & 56.4\\ 
            
            \uppercase\expandafter{\romannumeral3} & \Checkmark & \Checkmark &  \Checkmark & \Checkmark & & & 64.8 \\ 
            
            \uppercase\expandafter{\romannumeral4} & \Checkmark & \Checkmark &  \Checkmark & \Checkmark & \Checkmark & & 71.6 \\
            
            \uppercase\expandafter{\romannumeral5} & \Checkmark & \Checkmark  & \Checkmark & & \Checkmark & \Checkmark & 71.2 \\
            
            \uppercase\expandafter{\romannumeral6} & \Checkmark & & \Checkmark & \Checkmark & \Checkmark & \Checkmark & 70.5 \\
            
            \uppercase\expandafter{\romannumeral7} & & \Checkmark & \Checkmark & \Checkmark & \Checkmark & \Checkmark & 61.5 \\   
            
            \uppercase\expandafter{\romannumeral8} &\Checkmark & \Checkmark &  \Checkmark & \Checkmark & \Checkmark & \Checkmark & \textbf{72.4}\\   
            
            \bottomrule                                   
        \end{tabular}
        \end{footnotesize}
    }    
    \vspace{0.1cm}
    \caption{Ablation Study. Exp.\uppercase\expandafter{\romannumeral1} uses $\ell_2$ loss to align positive feature pairs. \textbf{ST}: Self-Training. \textbf{Proj}: Non-linear Projector $\Phi$. \textbf{Context}: Context-aware Consistency. \textbf{CL}: Vanilla Contrastive Loss. \textbf{Dir}: Directional Mask $\mathcal{M}^{h,w}_{d}$ defined in Eq.~\eqref{loss:dir_mask}. \textbf{NS}: Negative Sampling. \textbf{PF}: Positive Filtering.}
    \label{tab:ablation}   
\vspace{-0.3cm}
\end{table}

We conduct an extensive ablation study in Table~\ref{tab:ablation} to show the contribution of each component. The ablation study is based on 1/8 labeled data on PASCAL VOC. We use PSPNet with ResNet50 as the segmentation network. We set up two baselines, \ie, the model trained with only supervised loss (SupOnly) and the model with the self-training technique (ST). We follow~\cite{selftraining} to implement ST. 
We find that the results of three data lists do not vary much, so we only run on a single data list for ablation study.

\vspace{-0.3cm}
\paragraph{Context-aware Consistency}
To manifest the effectiveness of the proposed context-aware consistency, we make a comparison between the model with context-aware consistency and that only with low-level transformations. Intuitively, 
when the two patches cropped from an unlabeled image totally overlap with each other, \ie, the IoU is confined to [1, 1], there will be no contextual augmentation between them. In Table~\ref{tab:ablation}, the experiments \uppercase\expandafter{\romannumeral6} and \uppercase\expandafter{\romannumeral8} show that the model with our proposed context-aware consistency (Exp.\uppercase\expandafter{\romannumeral8}) is superior to that with only low-level transformations (Exp.\uppercase\expandafter{\romannumeral6}) by 1.9 points. To further highlight the improvement throughout the training process, we present the validation curves on PASCAL VOC and Cityscapes in Fig.~\ref{fig:contextual_vis}, where a huge gap can be observed.  

\vspace{-0.3cm}
\paragraph{Directional Contrastive Loss} The proposed DC Loss is a stronger constraint than $\ell_2$ loss, because $\ell_2$ loss does not consider pushing negative samples away. To illustrate this, in Table~\ref{tab:ablation}, we compare the models with $\ell_2$ loss (Exp.\uppercase\expandafter{\romannumeral1}) and DC Loss (Exp.\uppercase\expandafter{\romannumeral8}). It shows that simply using $\ell_2$ loss even worsens the performance from 65.0 to 64.2. Though the result of Exp.\uppercase\expandafter{\romannumeral3} without negative sampling 
also falls behind that of the baseline (SupOnly),
Exp.\uppercase\expandafter{\romannumeral4} shows that it is caused by the overwhelming false negative samples. After addressing the negative sampling problem
, the proposed DC Loss (Exp.\uppercase\expandafter{\romannumeral4}) surpasses the simple $\ell_2$ alignment (Exp.\uppercase\expandafter{\romannumeral1}) as well as the baseline (SupOnly) by a large margin.

Also, by comparing the experiments \uppercase\expandafter{\romannumeral5} and \uppercase\expandafter{\romannumeral8}, we observe that with the directional mask, the DC Loss improves the vanilla Contrastive Loss by 1.2 points. This demonstrates the effectiveness of the proposed directional alignment compared to the bilateral one.

\begin{table}
    \centering
    \tabcolsep=0.2cm
    {
        \begin{footnotesize}
        \begin{tabular}{c c c c c c c c}
            \toprule
            NumNeg
            & 500
            & 1k
            & 2k & 6.4k & 12.8k & 19.2k & 25.6k \\

            \specialrule{0em}{0pt}{2pt}
            \hline
            \specialrule{0em}{2pt}{0pt}
            
            mIoU & 70.9 & 71.0 & 71.7 & 71.3 & 71.9 & 72.4 & 71.9 \\ 
            
            \bottomrule                                   
        \end{tabular}
        \end{footnotesize}
    }
    \vspace{0.1cm}
\caption{Performance (mIoU) evaluated on PASCAL VOC under different number of negative samples per GPU.} 
\label{table:negsample}
\vspace{-0.3cm}
\end{table}

\vspace{-0.3cm}
\paragraph{Negative Sampling} \label{exp:ablation_ns} The experiments \uppercase\expandafter{\romannumeral3} and \uppercase\expandafter{\romannumeral4} in Table \ref{tab:ablation} show that the proposed negative sampling strategy with pseudo labels significantly improves the 
DC Loss. 

In Table~\ref{table:negsample}, we also notice that, within a certain scope, the more negative samples we use in training, the better performance we will get until it reaches an upper bound. More importantly, it is worth noting that the increased number of negative samples does not affect the training efficiency much by using the gradient checkpoint function provided in PyTorch~\cite{pytorch}. Specifically, by increasing negative samples from 500 to 19.2k, the training memory consumption merely increases by about 800M on each GPU, 
and the average training time of each iteration only increases from 0.96s to 1.18s. 
The implementation details are elaborated in the supplementary file.

\vspace{-0.3cm}
\paragraph{Positive Filtering} We show the effect of positive filtering in Table~\ref{tab:ablation}. The comparison between experiments  \uppercase\expandafter{\romannumeral4} and \uppercase\expandafter{\romannumeral8} demonstrates that the proposed positive filtering strategy makes further improvements by 0.8 points.

\vspace{-0.3cm}
\paragraph{The Role of Projector $\Phi$}
In Table~\ref{tab:ablation}, we compare the performance with and without the projector in experiment \uppercase\expandafter{\romannumeral7} and \uppercase\expandafter{\romannumeral8}, which shows the contribution of the projector.

\begin{table}[!t]
    \centering
    \tabcolsep=0.35cm
    {
        \begin{footnotesize}
        \begin{tabular}{ 
           l
           l
           l
           c  }
            \toprule
            Methods
            & Backbone
            & Semi
            & Weakly
            \\
            \specialrule{0em}{0pt}{2pt}
            \hline
            \specialrule{0em}{2pt}{0pt}
             WSSN~\cite{wssl} & VGG-16 & -
             & 64.6 \\                     
      
             GAIN~\cite{gain} & VGG-16 & -
             & 60.5 \\    
             
             MDC~\cite{imgsup1}& VGG-16 & -
             & 65.7 \\  
             
             DSRG~\cite{DSRG} & VGG-16 & -
             & 64.3 \\ 
             
             Souly \textit{et al.}~\cite{advsemi} & VGG-16 & 64.1
             & 65.8 \\               
             
             FickleNet~\cite{ficklenet} & ResNet-101 & -
             & 65.8 \\ 
             
             CCT~\cite{CCT} & ResNet-50 & 69.4
             & 73.2 \\  
             
            \specialrule{0em}{0pt}{1pt}
            \hline
            \specialrule{0em}{1pt}{0pt}
             Ours & VGG-16 & 68.7
             & 69.3 \\                
             
             CCT$^\ddagger$ &  ResNet-50 & 72.8 
             & 74.6\\               
             
             Ours & ResNet-50 & \textbf{74.5}
             & \textbf{76.1} \\              
             
            \bottomrule                                   
        \end{tabular}
        \end{footnotesize}
    }
\caption{Results with extra image-level annotations. CCT$^\ddagger$: Reproduced with the same setting as ours. Semi: Semi-supervised setting. 
Weakly: the setting with extra image-level labels.}  
\label{table:weakly}   
\vspace{-0.2cm} 
\end{table}

\subsection{Extension with Extra Image-level Annotations}
\label{sec:weakly}

Following \cite{CCT}, the original 1464 training images of PASCAL VOC~\cite{pascalvoc} are given the pixel-wise annotations and the rest of 9118 augmented images in SBD~\cite{pascalvocaug} are provided with image-level annotations.
 In Table~\ref{table:weakly}, our method reaches 76.1\% mIoU and surpasses CCT~\cite{CCT} by a large margin. Amazingly, it is 1.0 points higher than the model trained with the full pixel-level annotations, which demonstrates the generalization ability of the proposed method.
 

\subsection{Visual Comparison}
\label{sec:vis_comparision}
Fig.~\ref{fig:vis_comparison} presents the visual comparison with the SupOnly and current state-of-the-art methods. We observe that the results of our method are generally superior to others.

\section{Conclusion}
In this work, we focus on the semi-supervised semantic segmentation problem. In order to alleviate the problem of excessively using contexts and enhance self-awareness, we have presented the context-aware consistency, where we explicitly require features of the same identity but with different contexts to be consistent. In addition, we propose Directional Contrastive Loss to conduct the alignment. Also, two effective sampling strategies are put forward to make further improvements. Extensive experiments show our method achieves new state-of-the-art results, and also generalizes well with extra image-level annotations.

{\small
\bibliographystyle{ieee_fullname}
\bibliography{egbib}
}

\end{document}